\documentclass[letterpaper]{article} 
\usepackage{aaai25}  
\usepackage{times}  
\usepackage{helvet}  
\usepackage{courier}  
\usepackage[hyphens]{url}  
\usepackage{graphicx} 
\usepackage{natbib}  
\usepackage{caption} 
\usepackage{algorithm}
\usepackage{algorithmic}

\usepackage{booktabs}
\usepackage{makecell}
\usepackage{comment}
\usepackage{multirow}
\usepackage{amsmath}
\usepackage{float} 
\usepackage{subcaption}
\def\etal{\emph{et al}.}
\usepackage{xcolor}
\usepackage{colortbl}
\usepackage{amssymb}
\usepackage{enumerate}
\usepackage{nicefrac}
\DeclareMathAlphabet{\mathbbm}{U}{bbold}{m}{n}

\usepackage{newfloat}
\usepackage{listings}
\DeclareCaptionStyle{ruled}{labelfont=normalfont,labelsep=colon,strut=off} 
\floatstyle{ruled}
\newfloat{listing}{tb}{lst}{}
\floatname{listing}{Listing}
\title{Point Cloud Semantic Segmentation with Sparse and Inhomogeneous Annotations}
\author {
    Zhiyi Pan\textsuperscript{\rm 1, \rm 2},
    Nan Zhang\textsuperscript{\rm 1},
    Wei Gao\textsuperscript{\rm 1}\thanks{Wei Gao is the corresponding author.},
    Shan Liu\textsuperscript{\rm 3},
    Ge Li\textsuperscript{\rm 1}
}
\affiliations {
    \textsuperscript{\rm 1}Guangdong Provincial Key Laboratory of Ultra High Definition Immersive Media Technology,\\Shenzhen Graduate School, Peking University\\
    \textsuperscript{\rm 2}Peng Cheng Laboratory\\
    \textsuperscript{\rm 3}Media Lab, Tencent\\ 
    \{panzhiyi, zhangnan\}@stu.pku.edu.cn, gaowei262@pku.edu.cn, shanl@tencent.com, geli@ece.pku.edu.cn
}

\usepackage{bibentry}

\begin{document}

\maketitle

\begin{abstract}
Utilizing uniformly distributed sparse annotations, weakly supervised learning alleviates the heavy reliance on fine-grained annotations in point cloud semantic segmentation tasks.
However, few works discuss the inhomogeneity of sparse annotations, albeit it is common in real-world scenarios.
Therefore, this work introduces the probability density function into the gradient sampling approximation method to qualitatively analyze the impact of annotation sparsity and inhomogeneity under weakly supervised learning.
Based on our analysis, we propose an \textbf{A}daptive \textbf{A}nnotation \textbf{D}istribution \textbf{Net}work (AADNet) capable of robust learning on arbitrarily distributed sparse annotations. Specifically, we propose a label-aware point cloud downsampling strategy to increase the proportion of annotations involved in the training stage. Furthermore, we design the multiplicative dynamic entropy as the gradient calibration function to mitigate the gradient bias caused by non-uniformly distributed sparse annotations and explicitly reduce the epistemic uncertainty. Without any prior restrictions and additional information, our proposed method achieves comprehensive performance improvements at multiple label rates and different annotation distributions.

\end{abstract}

\begin{links}
    \link{Code}{https://github.com/panzhiyi/AADNet}
\end{links}

\section{Introduction}
\begin{figure}[htbp]
	\centering
		\includegraphics[width=\linewidth]{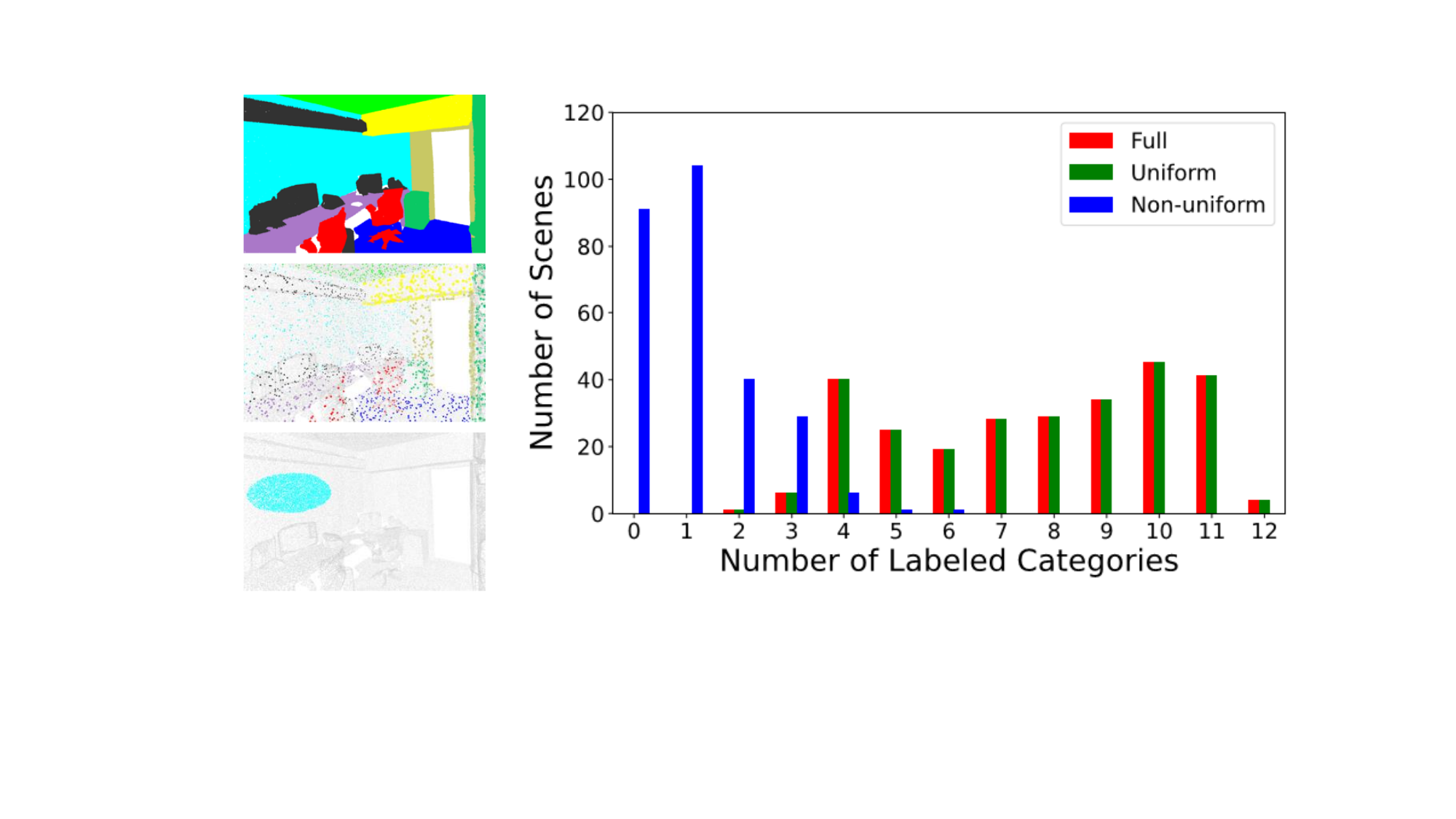}
	\caption{Visual comparison about full annotations, uniform sparse annotations, and non-uniform sparse annotations. Based on the data presented in regarding the number of labeled categories per scene, the distribution disparity between non-uniform sparse annotation and full annotation is pronounced.}
	\label{fig:main}
\end{figure}

Point cloud has become one of the most popular 3D representations due to its flexible ability to represent 3D objects and the wide deployment of capture devices. Point cloud semantic segmentation plays a crucial role in various applications~\cite{hu2023planning,cui2021deep} by exploiting the abundant 3D geometric information from point clouds. However, point cloud semantic segmentation suffers from heavy reliance on fine-grained annotated scenes~\cite{xu2020weakly}. The increase in data dimensionality brings an order of magnitude increase in structural complexity and processing difficulty, further exacerbating the annotation burden. Therefore, weakly supervised point cloud semantic segmentation is a promising topic for deployment and applications.

Existing weakly supervised methods rely on simulating sparse annotations through uniformly distributed sampling on a fully labeled dataset. We refer to such annotations as \textit{uniformly distributed sparse annotations}. The simulated dataset maintains distribution consistency between sparse and full annotations through uniformly random sampling~\cite{xu2020weakly} or "One-Thing-One-Click"~\cite{liu2021one}. However, it places strict constraints on the annotation distribution. 

\textit{Non-uniformly distributed sparse annotations} are inevitable in real-world scenarios, which can be illustrated by the following processes in weakly supervised learning:
\begin{itemize}
    \item The projection relationship between the 3D point cloud and the 2D annotation plane causes the non-uniformly distributed points on the annotation plane, directly leading to inhomogeneous annotations. Moreover, to improve annotation efficiency, annotators are more likely to annotate categories that are easier to label, which exacerbates the inhomogeneity.
    \item Supervoxels are typically used to expand sparse annotation information, during which uniformly distributed sparse annotated points are transformed into non-uniformly distributed sparse annotated supervoxels.
    \item Self-training in weakly supervised learning treats high-confidence predictions as pseudo-labels for subsequent network learning. Since high-confidence points are usually concentrated in local regions with significant patterns, the pseudo-labels are also unevenly distributed.
\end{itemize}

As illustrated in Fig.~\ref{fig:main}, there is a significant distribution gap between non-uniformly distributed sparse annotations and full annotations. Without the uniform sampling prior, the inhomogeneous annotations impede weakly supervised learning. The current approaches fail to consider the essential contribution of sparse labeling inhomogeneity to point cloud segmentation. Therefore, our work extends the annotation requirements from uniformly distributed sparse annotations to arbitrarily distributed sparse annotations.


To investigate the impact of annotations, we introduce the probability density function into the gradient sampling approximation analysis~\cite{xu2020weakly}. According to the Central Limit Theorem, the gradient discrepancy between weak supervision and full supervision follows a normal distribution, where the sparsity and inhomogeneity of labeling affect its mean and variance, respectively. Based on this analysis, we propose an \textbf{A}daptive \textbf{A}nnotation \textbf{D}istribution \textbf{Net}work (AADNet) to learning from arbitrarily distributed annotations, which is composed of label-aware downsampling strategy and multiplicative dynamic entropy.


Downsampling is essential for point cloud scenes due to computational limitations. However, random downsampling~\cite{hu2020randla} directly borrowed from the full supervision would miss valuable annotations entailed for training. Conversely, sampling only labeled points would severely sacrifice the structural information from point cloud resampling. Therefore, we propose the label-aware point cloud downsampling strategy (LaDS) to increase the proportion of labeled points involved in the training stage and retain rich structured information to a great extent. 


We propose the multiplicative dynamic entropy with asynchronous training (MDE-AT) to accommodate the annotation inhomogeneity. We prove that an ideal calibration function for correcting the gradient bias should be inversely proportional to the probability density. However, the probability density of the annotation distribution is agnostic. Therefore, we utilize multiplicative dynamic entropy as an alternative calibration function. Additionally, to improve the epistemic certainty~\cite{kendall2017uncertainties}, we employ asynchronous training that iteratively imposes entropy loss and partial cross-entropy loss with the calibration function.

The contributions of this work can be summarized in the following three aspects:
\begin{itemize}
    \item To the best of our knowledge, we are the first to discuss the inhomogeneity of sparse annotations in weakly supervised learning. The quantitative impact of non-uniformly distributed sparse annotations on network learning is revealed through gradient sampling approximation analysis.
    \item Based on the analysis, we propose an Adaptive Annotation Distribution Network that can handle sparsely annotated data with arbitrary distributions.
    \item The effectiveness of AADNet is comprehensively validated on multiple annotation settings and datasets.
\end{itemize}

\section{Related Work}
For weakly supervised point cloud semantic segmentation, researchers have utilized well-established techniques in 2D vision, such as class activation maps~\cite{zhou2016learning}, perturbation consistency~\cite{pan2021scribble}, label propagation~\cite{iscen2019label}, and contrastive learning~\cite{chen2020simple}. These techniques leverage rich hypothetical priors, processing paradigms, or additional supervision to enable weakly supervised point cloud semantic segmentation. Rather than relying on a single technique, most methods build their effective weak supervision frameworks by integrating multiple techniques.

\noindent \textbf{Class Activation Maps.} In weakly supervised point cloud semantic segmentation, Class Activation Maps (CAMs) provide category localization information from both scene-level and sub-scene-level annotations. To further improve the semantic localization capability of CAMs, MPRM~\cite{wei2020multi} and Song~\etal~\cite{song2022learning} act CAMs and geometric projection to obtain point-level pseudo-labels. MILTrans~\cite{yang2022mil} utilizes an adaptive global weighted pooling mechanism on CAMs to mitigate the negative effects of irrelevant classes and noisy points.

\noindent \textbf{Perturbation Consistency.} Consistency between the corresponding features of the original and perturbed point clouds provides additional supervision signals. Xu~\etal~\cite{xu2020weakly} and PSD~\cite{zhang2021perturbed} apply perturbations such as rotation and mirror flipping to the point cloud scene, while MILTrans~\cite{yang2022mil} maintains consistency on the pair point clouds before and after downsampling. DAT~\cite{wu2022dual} and HybridCR~\cite{li2022hybridcr} dynamically construct point cloud replicas using adaptive gradient and embedding network learning, respectively. RPSC~\cite{lan2023weakly} and PointMatch~\cite{wu2022pointmatch} leverage pseudo-labels to convey consistent information.

\noindent \textbf{Label Propagation.} The label propagation generates high-quality pseudo-labels in self-training. For instance, OTOC~\cite{liu2021one} and OTOC++~\cite{liu2023one} propose RelationNets to accurately measure the similarity between 3D graph nodes and propagate labeling information. Zhang~\etal~\cite{zhang2021weakly} develops a sparse label propagation algorithm guided by the category prototype~\cite{snell2017prototypical}. 

\noindent \textbf{Contrastive Learning.} Contrastive learning selects anchor points and imposes restrictions on corresponding positive point pairs and negative point pairs. HybridCR~\cite{li2022hybridcr} constructs contrastive learning on transformed point cloud pairs, local geometry pairs, and category prototype pairs, while MILTrans~\cite{yang2022mil} performs contrastive loss between category pairs in the scene. Besides, pre-training methods with contrastive learning~\cite{xie2020pointcontrast,hou2021exploring} can also achieve semantic segmentation tasks under few annotations.

\begin{figure*}[t]
    \centering
    \includegraphics[width=\linewidth]{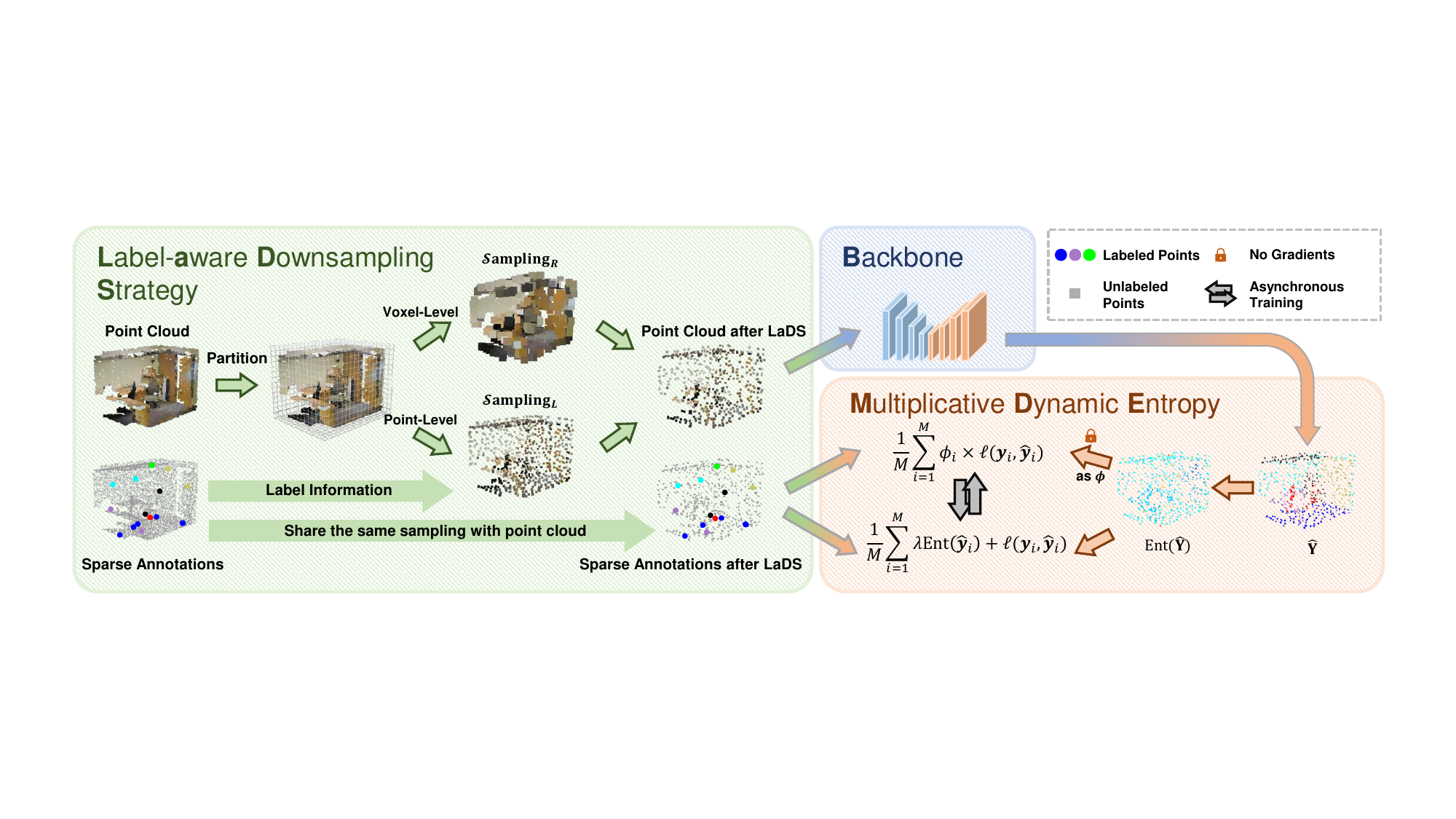}
    \caption{The framework of AADNet consists of two modules: (1) Label-aware downsampling strategy (LaDS) to boost the training annotation rate and maintain structural diversity. (2) Multiplicative dynamic entropy with asynchronous training (MDE-AT) to correct for gradient bias due to sparse labeling distributions as well as to explicitly improve epistemic certainty.}
    \label{fig:network}
\end{figure*}

\noindent \textbf{Others.} SQN~\cite{hu2022sqn} collects a group of hierarchical representations within the locally labeled neighborhoods through interpolation. LessIsMore~\cite{pan2024less} proposes a sparse labeling recommendation framework for weakly supervised learning. REAL~\cite{kweon2024weakly} introduces 2D information from SAM~\cite{kirillov2023segment} to weak supervision. DGNet~\cite{pan2024distribution} explicitly aligns the feature space and prior distribution to provide additional supervised signals.

The above methods are verified under the uniformly distributed sparse annotations. However, uniformly distributed sampling is elusive to achieve. Consequently, we propose a point cloud semantic segmentation method with adaptive annotation distribution for arbitrary sparse annotation scenarios.

\section{Method}
In Sec.~\ref{sec:overview}, we give a brief overview of weakly supervised point cloud semantic segmentation. The gradient sampling approximation analysis with probability density function is introduced to investigate the annotation distribution in Sec.~\ref{sec:Mehthdology}. Based on the analysis, we propose the Adaptive Annotation Distribution Network, which is composed of a label-aware downsampling strategy (Sec.~\ref{sec:LaDS}) and multiplicative dynamic entropy (Sec.~\ref{sec:MDE}). The framework of AADNet is illustrated in Fig.~\ref{fig:network}.

\subsection{Preliminaries} \label{sec:overview}
A point cloud scene in the training set for semantic segmentation can be denoted as $\mathbf{P}=\{\mathbf{X}, \mathbf{Y}\}$, where $\mathbf{X}$ and $\mathbf{Y}$ represent point set and its corresponding label, respectively. In detail, $\mathbf{X} = [\mathbf{x}_1,\mathbf{x}_2,\dots,\mathbf{x}_N]$ and $\mathbf{x}_i \in \mathbb{R}^{1\times F}$, in which $N$ denotes the number of points and $F$ is the dimension number of initial feature on each point $\mathbf{x}_i$. Without losing generality, $\mathbf{Y}=[\mathbf{y}_1,\mathbf{y}_2,\dots,\mathbf{y}_M]$ and $\mathbf{y}_i\in \{0,1\}^{1\times C}$, where $M$ and $C$ indicate the number of labeled points and categories, respectively. When $N \gg M$, the task is termed as a weakly supervised point cloud semantic segmentation task, and the label rate is $\frac{M}{N}$. The partial cross-entropy loss, which is commonly used in weak supervision, is defined as:
\begin{equation}
    \mathcal{L}_p = \frac{1}{M}\sum_{i=1}^M \ell (\mathbf{y}_i, \hat{\mathbf{y}}_i) = \frac{1}{M}\sum_{i=1}^M \mathbf{y}_{i} {\rm log} (\hat{\mathbf{y}}_{i}^\top),
\end{equation}
where $\hat{\mathbf{y}}_i \in [0,1]^{1\times C}$ denotes the prediction probability at point $\mathbf{x}_i$.

\subsection{Gradient Sampling Approximation Analysis} \label{sec:Mehthdology}
For two networks with the same structure and initialization parameters, it is assumed that the closer the gradients are, the more likely they will converge to the same prediction results for the same input~\cite{xu2020weakly}. The gradients of cross-entropy $\nabla \mathcal{L}$ and partial cross-entropy $\nabla \mathcal{L}_p$ are formulated as:
\begin{equation}
\begin{aligned}
    \nabla \mathcal{L} &= \frac{1}{N} \sum_{i=1}^N g(\hat{\mathbf{y}}_i),\\
    \nabla \mathcal{L}_p &= \frac{1}{M} \sum_{i=1}^M g(\hat{\mathbf{y}}_i),
\end{aligned}
\end{equation}
where $g(\hat{\mathbf{y}}_i)=\nabla \ell (\mathbf{y}_i, \hat{\mathbf{y}}_i)$, denotes the loss gradient at $\mathbf{x}_i$.  Weakly supervised learning can be viewed as fully supervised learning with $M$ times sampling of gradients, and the average gradient value of $M$ times sampling is equal to $\nabla \mathcal{L}_p$. Assuming that $g(\hat{\mathbf{y}}_i)$ satisfies the independent identical distribution, the uniformly distributed sampled $s(g(\hat{\mathbf{y}}_i))$ follows the distribution $\mathcal{D}_u(E_u,V_u)$. The expectation $E_u = \mathbb{E}_{\mathbf{x}_i \sim \mathcal{D}}[p_i g(\hat{\mathbf{y}}_i)]=\sum_i \frac{1}{N} g(\hat{\mathbf{y}}_i)=\nabla \mathcal{L}$, where $\mathcal{D}$ is the distribution of points and $p_i=\frac{1}{N}$ denotes the probability density function of uniformly distributed sampling. According to the Central Limit Theorem, the following convergent distribution can be obtained under uniformly distributed sampling:
\begin{equation}
\begin{aligned}
    (\nabla \mathcal{L}_p - \nabla \mathcal{L}) \sim \mathcal{N}(0,\frac{V_u}{M}).
\end{aligned}
\label{eq:uniform_centrallimit}
\end{equation}
Eq.~\ref{eq:uniform_centrallimit} illustrates that the gradient difference between full annotations and uniformly sparse annotations obeys a normal distribution with $0$ expectation and $\frac{V_u}{M}$ variance, which indicates that the average gradient of uniform sparse annotations is more likely to be similar to that of full annotations when the label rate becomes larger.

Based on the previous assumption, the prediction results of the model supervised by uniform sparse annotations at an appropriate label rate are consistent with those of the fully supervised semantic segmentation model. However, this corollary is not universal for the non-uniformly distributed sparse annotations. We assume that non-uniformly sampled gradient $s_n(g(\hat{\mathbf{y}}_i))$ follows the distribution $\mathcal{D}_n(E_n,V_n)$. The expectation $E_n = \mathbb{E}_{\mathbf{x}_i \sim \mathcal{D}}[p'_i g(\hat{\mathbf{y}}_i)]=\sum_i p'_i g(\hat{\mathbf{y}}_i)=\nabla \mathcal{L} + {\rm \Delta}$, where $p'_i$ denotes the probability density function of the non-uniformly sampling distribution and the bias ${\rm \Delta}=\sum_i (p'_i-p_i) g(\hat{\mathbf{y}}_i)$. Similarly, according to the Central Limit Theorem, we obtain the following convergent distribution under non-uniformly distributed sampling:
\begin{equation}
\begin{aligned}
    (\nabla \mathcal{L}_p - \nabla \mathcal{L}) \sim \mathcal{N}({\rm \Delta} ,\frac{V_n}{M}).
\end{aligned}
\label{eq:nonuniform_centrallimit}
\end{equation}
Eq.~\ref{eq:nonuniform_centrallimit} demonstrates that in the case of non-uniformly distributed sparse annotations, the average gradient still differs by ${\rm \Delta}$ even at large label rates. We conclude that \textbf{the performance of weak supervision is not only affected by the annotation sparsity but also the annotation inhomogeneity}. Without any priors, uniformly distributed sparse labeling is the optimal sparse labeling.


\subsection{Label-aware Downsampling Strategy} \label{sec:LaDS}
The downsampling strategy determines the upper bound of the receptive field and the diversity of the structural information of the scene~\cite{hu2020randla}. However, the random downsampling strategy used by full supervision is not applicable to sparse labeling scenarios. The random downsampling treats labeled and unlabeled points equally, significantly weakening the semantic supervision information of the sampled point cloud. An intuitive alternative is preferentially selecting all labeled points. Although it can greatly preserve labeled points after downsampling, it compromises the diversity of structural information when sampling the point cloud repeatedly. Therefore, a label-aware downsampling strategy is proposed to preserve labeled points and abundant structural topology.

\begin{figure}[tbp]
    \centering
    \includegraphics[width=\linewidth]{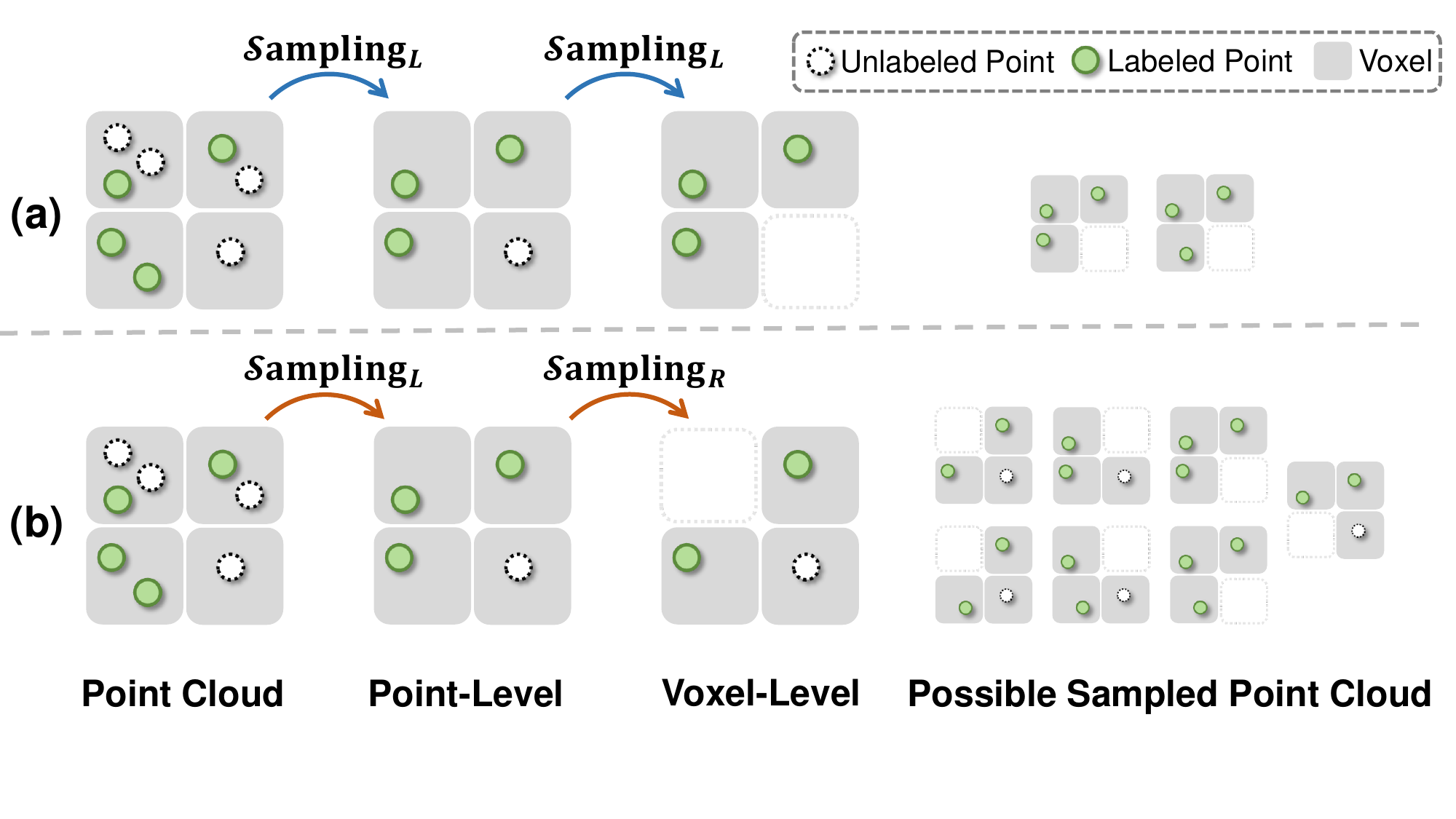}
    \caption{Compared with (a) voxel-level label-first downsampling, (b) voxel-level random downsampling retains diverse scene structural information.}
    \label{fig:LaDS}
\end{figure}

The label-aware point cloud downsampling strategy is shown in Fig.~\ref{fig:network}. First, we partition the entire point cloud scene by voxels, \textit{i.e.}, $\mathcal{V}=\{\mathbf{v}_j\}$ and $\mathbf{v}_{j_1}\cap \mathbf{v}_{j_2} = \varnothing$ when $j_1 \neq j_2$. Based on the voxel partition, the downsampling on the point cloud decomposes into point-level label-first downsampling $\mathcal{S}\text{ampling}_L$ and voxel-level random downsampling $\mathcal{S}\text{ampling}_R$.

\begin{enumerate}[1)]
    \item \textbf{Point level label-first downsampling $\mathcal{S}\text{ampling}_L$}. For the voxel that contains labeled points, one labeled point is randomly selected among the labeled points within the voxel, and for the voxel that does not contain any labeled point, one point is randomly selected within the voxel. The loss of structural diversity with point-level label-first sampling is almost negligible since the sampling is over a small range of voxels.
    \item \textbf{Voxel-level random downsampling $\mathcal{S}\text{ampling}_R$}. At the voxel level, we retain the conventional random downsampling strategy. In contrast to voxel-level label-first downsampling, voxel-level random downsampling sacrifices a small percentage of the labeled points but retains diverse scene structural information when sampling repeatedly in different training epochs. A comparison of sampling strategies with voxel-level label-first downsampling and voxel-level random downsampling is visualized in Fig.~\ref{fig:LaDS}.
\end{enumerate}

The points $\mathbf{P}'$ sampled by the label-aware point cloud downsampling strategy can be defined as:

\begin{equation}
    \mathbf{P}' = \mathcal{S}\text{ampling}_R \bigg( \{ \mathcal{S}\text{ampling}_L (\mathbf{v}_j,1)\ |\ \mathbf{v}_j \in \mathcal{V} \},K \bigg),
\end{equation}
where $K$ is a predefined number of sampling. $\mathcal{S}\text{ampling}(\mathbf{v},t)$ denotes sample $t$ times from the set $\mathbf{v}$.

It is worth noting that the label-aware downsampling strategy targets annotation sparsity and imposes no restrictions on annotation inhomogeneity.



\subsection{Multiplicative Dynamic Entropy} \label{sec:MDE}
\begin{figure}[t]
    \centering
    \includegraphics[width=0.9\linewidth]{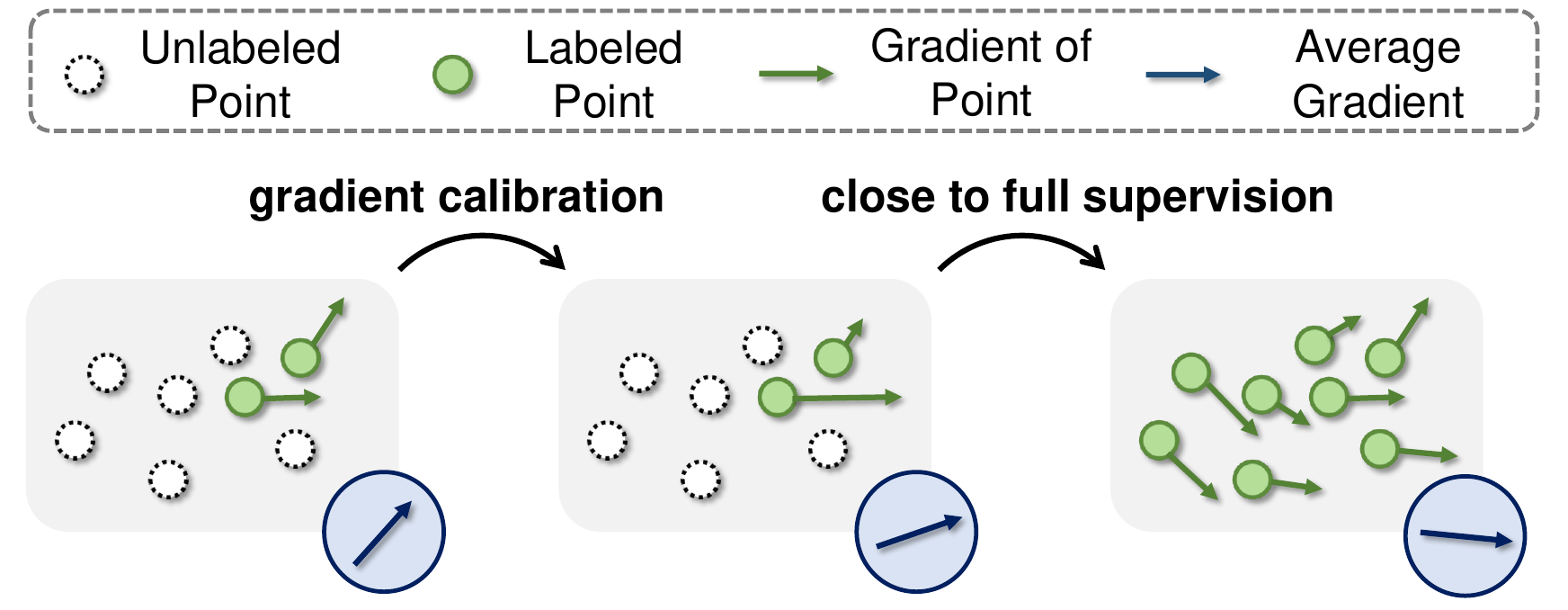}
    \caption{Schematic diagram of gradient calibration by MDE.}
    \label{fig:MDE}
\end{figure}

Based on the methodological analysis in Sec.~\ref{sec:Mehthdology}, we impose a gradient calibration function $\phi$ on $\mathcal{L}_p$ to counteract the gradient expectation bias caused by the non-uniformly distributed sparse annotations. Ideally, using multiplicative calibration function $\phi_i=\frac{p_i}{p'_i}$, the expectation of the corrected gradient satisfies:
\begin{equation}
    E_n=\mathbb{E}_{\mathbf{x}_i \sim \mathcal{D}}[p'_i \phi_i g(\hat{\mathbf{y}}_i)]=\mathbb{E}_{\mathbf{x}_i \sim \mathcal{D}}[p_i g(\hat{\mathbf{y}}_i)]=\nabla \mathcal{L}.
\end{equation}
As shown in Fig.~\ref{fig:MDE}, the gradient bias ${\rm \Delta}$ is corrected by $\phi$. Compared to the ideal additive calibration function $\psi_i=(p'_i-p_i) g(\hat{\mathbf{y}}_i)$, the multiplicative calibration function $\phi$ is gradient-independent and easier to estimate.

Nevertheless, the probability density function $p'_i$ of non-uniformly distributed sampling in the ideal $\phi$ is agnostic. Taking the negative correlation between labeling difficulty and labeling probability $p'_i$ as a guideline, we utilize the entropy function as the dynamic calibration function, \emph{i.e.}, $\phi_i=\mathrm{Ent}(\hat{\mathbf{y}}_i)= \hat{\mathbf{y}}_{i} {\rm log} (\hat{\mathbf{y}}_{i}^\top)$. Shannon entropy as an alternative multiplicative calibration function offers two major benefits:
\begin{enumerate}[1)]
    \item \textbf{Adaptive.} Entropy is a dynamic function related to sparsely labeled scenes in the training stage, which can adaptively correct for different point cloud scenes.
    \item \textbf{Representativity.} The classification difficulty of points reflects the annotation distribution. Since entropy possesses the ability to reflect classification difficulty, it also represents annotation distribution.
\end{enumerate}
It is worth noting that although $\phi$ is a function on predictions $\{\hat{\mathbf{y}}_i \}$, it does not generate gradients during training, but only serves to weight the partial cross-entropy loss dynamically.

To make entropy more accurate and reduce the epistemic uncertainty of the network, we conversely constrain its learning as a loss term. Entropy loss and partial cross-entropy loss are organized in an asynchronous training manner. The reason for introducing asynchronous training is that synchronous training will change the optimization objective of the network and prevent partial cross-entropy loss from converging. The total loss function $\mathcal{L}^k_{\mathrm{AAD}}$ at the $k$-th training epoch can be defined as:
\begin{equation}
\begin{aligned}
    \mathcal{L}^k_{\mathrm{AAD}}=&\mathbbm{1}(k \bmod 2\tau \geq \tau)\frac{1}{M}\sum_{i=1}^M \Big [ \phi_i \times \ell (\mathbf{y}_i, \hat{\mathbf{y}}_i)\Big ] + \\
    &\mathbbm{1}(k\bmod 2\tau < \tau) \frac{1}{M}\sum_{i=1}^M \Big [ \lambda\mathrm{Ent}(\hat{\mathbf{y}}_i) + \ell (\mathbf{y}_i, \hat{\mathbf{y}}_i)\Big ],
\end{aligned}
\label{eq:lossfunction}
\end{equation}
where $\mathbbm{1}(\cdot)$ denotes the indicator function, $\lambda$ is a predefined balance weight, and $\tau$ denotes the step interval of asynchronous training. To learn efficiently, AADNet still imposes partial cross-entropy loss while training via $\mathrm{Ent}(\hat{\mathbf{y}}_i)$.

The gradient calibration function targets annotation inhomogeneity and has no restriction on annotation sparsity. It is worth noting that even under uniformly sampled sparse annotations, due to extreme sparsity, the distribution tends to exhibit inhomogeneity, allowing the gradient calibration function to play a positive role.

\section{Labeling Implementation}

The quality of weakly supervised annotations depends on the sparsity and inhomogeneity of the annotations. Annotation sparsity can be easily controlled by adjusting the label rate.\footnote{Compared with "One-Thing-One-Click", the uniformly random sampling allows continuous adjustment of the label rate.} In this section, we introduce how to control the degree of non-uniformity through simulation, which is essential for a comprehensive evaluation.

Consider the most uneven case of sparse annotations, where all annotations are concentrated in a highly similar local region, effectively meaning that only one point cloud cluster in the scene is labeled. As the number of labeled point cloud clusters increases (with the scale of each cluster decreasing to maintain the same label rate), the homogeneity of the sparse annotations improves. When the number of annotated point cloud clusters is further increased to $M$, with each cluster containing only one point, the sparse annotation becomes equivalent to uniform sampling across annotations of all points. Therefore, controlling the number of point cloud clusters can effectively simulate the annotation inhomogeneity.

\begin{figure}[tbp]
    \centering
    \includegraphics[width=0.9\linewidth]{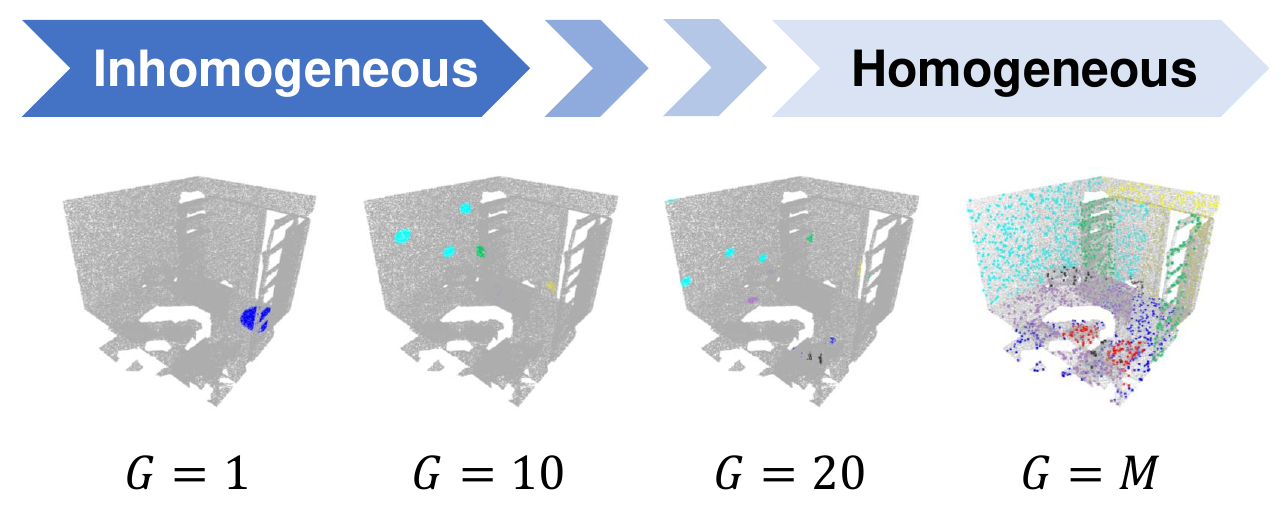}
    \caption{The sparse annotations with different $G$.}
    \label{fig:inhomo}
\end{figure}

In our annotation implementation, under the condition of a label rate of $\frac{M}{N}$, we first define the number of point cloud clusters $G$, then randomly sample $G$ points in the scene as the cluster centers. Based on the initial features of the points, we then query $\lfloor \frac{M}{G} \rfloor$ points for each cluster center as the annotated point cloud clusters. Fig.~\ref{fig:inhomo} illustrates the simulated sparse annotations with different cluster numbers.

\section{Experiments}
\subsection{Experiment Setting}
\textbf{Dataset.} S3DIS~\cite{armeni20163d} contains 271 rooms in six areas with 13 semantic categories. We train our model on Area 1, 2, 3, 4, and 6, and evaluate the segmentation performance on Area 5. ScanNetV2~\cite{dai2017scannet} consists of 1,513 scanned scenes obtained from 707 different indoor environments and provides 21 semantic categories for each point. We utilize 1,201 scenes for training and 312 scenes for validation, as per the official split. The outdoor dataset SemanticKITTI~\cite{behley2019semantickitti} with 19 classes is also considered. Point cloud sequences 00 to 10 are used in training, with sequence 08 as the validation set. 

\noindent \textbf{Annotation setting.} To compare with alternatives, the performance at label rates of 0.01\% on S3DIS, 0.1\% on SemanticKITTI, and 20 points per scene (label rate is about 0.014\%) on ScanNetV2 are reported. For full validation, we set $G$ to 1, 10, 20, and $M$ to regulate the inhomogeneity of sparse labeling.

\noindent \textbf{Implementation.} We take RandLA-Net~\cite{hu2020randla} and PointNeXt-L~\cite{qian2022pointnext} as the backbones to construct AADNet. Unless otherwise noted, AADNet is trained with default settings. To prevent the entropy loss from misleading the network during the early stage of training~\cite{pan2024cc4s}, we set the start epoch of asynchronous training to 50. The weight $\lambda=0.01$ and step interval $\tau=5$ in Eq.~\ref{eq:lossfunction}. Our models are trained with one NVIDIA V100 GPU on S3DIS, eight NVIDIA TESLA T4 GPUs on ScanNetV2, and four NVIDIA V100 GPUs on SemanticKITTI.

\begin{figure*}[htbp]
	\centering
		\includegraphics[width=0.82\linewidth]{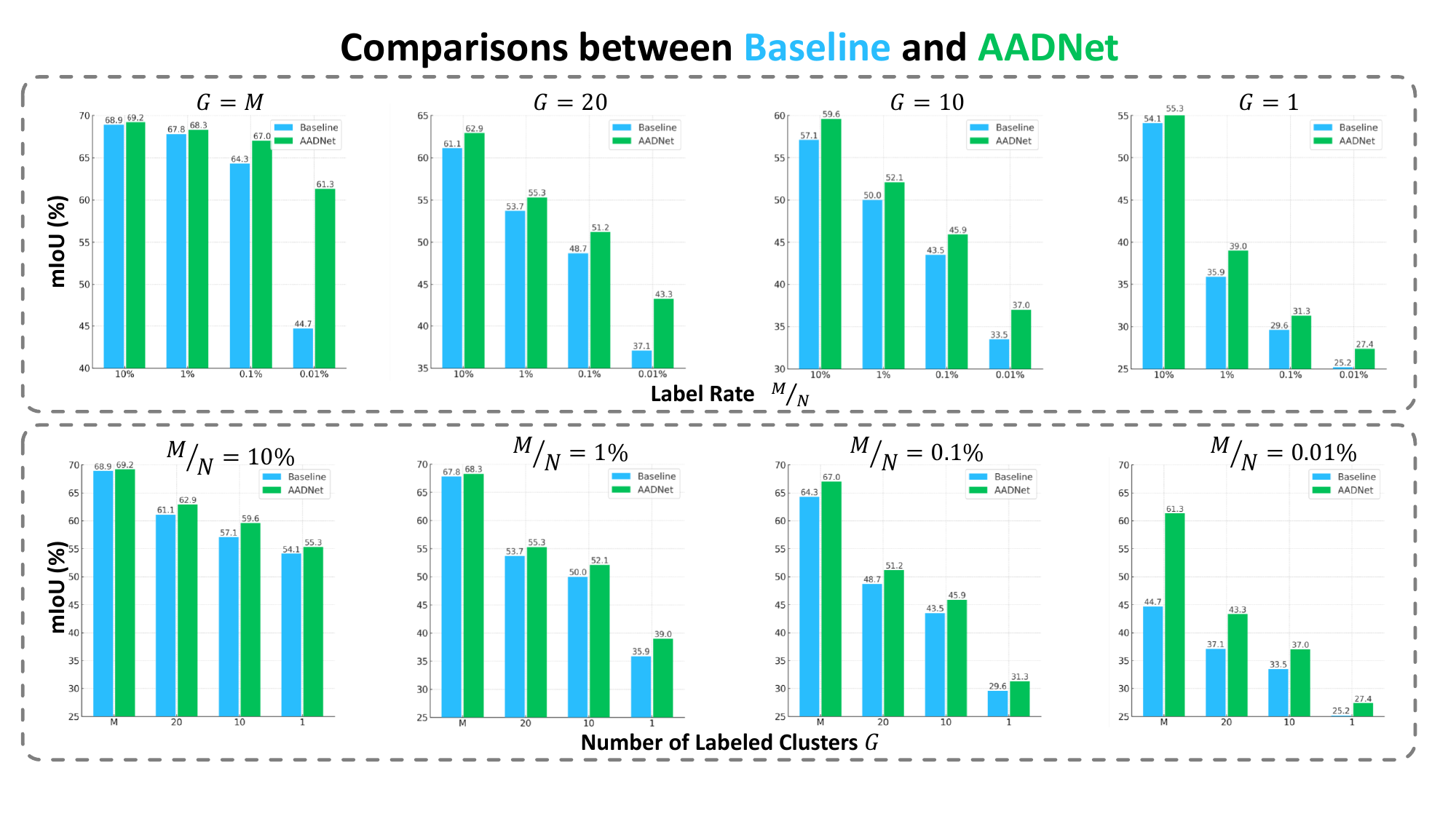}
	\caption{Comparisons between Baseline (PointNeXt) and AADNet with different label rates and numbers of labeled clusters.}
	\label{fig:baseline_inhomo}
\end{figure*}

\subsection{Comparison Results}

\begin{table}[t]
\begin{center}
\resizebox{\linewidth}{!}{
\setlength{\tabcolsep}{8.7pt}
\begin{tabular}{l|c|c|c}
\toprule[3\arrayrulewidth]
Method & $\nicefrac{M}{N}$ & S3DIS & ScanNetV2 \\
\specialrule{0em}{1pt}{1pt} \hline
\specialrule{0em}{1pt}{1pt}
SQN~\cite{hu2022sqn}                                         & 10\%  & 38.4                &   41.1                   \\
EDRA~\cite{tang2024all}& 10\%                                           & 43.3                         &   46.2                 \\
RandLA-Net~\cite{hu2020randla} & 10\%                                          & 40.9                &   42.9                   \\
\rowcolor{green!10} \textbf{AADNet (RandLA-Net)}  & 10\%                                         & \textbf{50.5}               &   \textbf{53.0}                  \\
\hline
DCL~\cite{yao2023weakly}  & 10\%                                         & 52.9                     &   55.1                        \\
CPCM~\cite{liu2023cpcm}  & 10\%                                        & 54.9                       & 59.5                         \\
PointNeXt~\cite{qian2022pointnext}  & 10\%                                    & 54.1                   & 56.0                   \\
\rowcolor{green!10} \textbf{AADNet (PointNeXt)}     & 10\%                                   & \textbf{56.0}                  & \textbf{61.2}                 \\
\specialrule{0em}{1pt}{1pt} \hline
\specialrule{0em}{1pt}{1pt}

SQN~\cite{hu2022sqn}                                         & 1\%  & 30.7                 &  32.6                        \\
EDRA~\cite{tang2024all}& 1\%                                           & 34.6                 &  35.7                          \\
RandLA-Net~\cite{hu2020randla} & 1\%                                          & 34.1                     &   36.4              \\
\rowcolor{green!10} \textbf{AADNet (RandLA-Net)}  & 1\%                                         & \textbf{37.7}           &   \textbf{41.1}                 \\
\hline
DCL~\cite{yao2023weakly}  & 1\%                                             &  35.6                   &  34.7                 \\
CPCM~\cite{liu2023cpcm}  & 1\%                                    &  38.2                    &  41.7                 \\
PointNeXt~\cite{qian2022pointnext}  & 1\%                           & 35.9               & 37.0              \\
\rowcolor{green!10} \textbf{AADNet (PointNeXt)}     & 1\%                                        & \textbf{39.0}                 & \textbf{43.4}              \\
\specialrule{0em}{1pt}{1pt}
\toprule[3\arrayrulewidth]  
\end{tabular}
}
\end{center}
\caption{Quantitative comparisons with inhomogeneous annotations and $G=1$.}
\label{tab:non-uniform}
\end{table}

\begin{table}[t]
\begin{center}
\resizebox{\linewidth}{!}{
\begin{tabular}{l|cc|cc|cc}
\toprule[3\arrayrulewidth]

\multicolumn{1}{l|}{\multirow{2}{*}{Method}} & \multicolumn{2}{c|}{S3DIS} & \multicolumn{2}{c|}{ScanNetV2} & \multicolumn{2}{c}{SemanticKITTI}\\
\multicolumn{1}{c|}{}                        &  $\nicefrac{M}{N}$         &    mIoU (\%)      &  $\nicefrac{M}{N}$         &    mIoU (\%) &  $\nicefrac{M}{N}$         &    mIoU (\%)        \\
\specialrule{0em}{1pt}{1pt} \hline
\specialrule{0em}{1pt}{1pt}
SQN~\cite{hu2022sqn}                                      & 0.01\%  & 45.3               &   - &   - & 0.1\% & 50.8                   \\
CPCM~\cite{liu2023cpcm}                                      & 0.01\%  & 49.3               &   0.01\% &   52.2 & 0.1\% & 44.0                   \\
EDRA~\cite{tang2024all}   & 0.02\%  & 48.4               &   20pts &   57.0 & - & -          \\
RPSC~\cite{lan2023weakly} & 0.04\%  & 64.0               &   50pts &   58.7 & 0.1\% & 50.9          \\
PointMatch~\cite{wu2022pointmatch} & 0.01\%  & 59.9               &   20pts &   62.4 & - & -          \\
MILTrans~\cite{yang2022mil} & 0.02\% & 51.4 & 20pts & 54.4 & - & - \\
\hline
RandLA-Net~\cite{hu2020randla} & 0.01\%  &  38.6 & 20pts & 50.7 & 0.1\% & 43.1  \\
\rowcolor{green!10} \textbf{AADNet (RandLA-Net)}  & 0.01\%  &  \textbf{55.6} & 20pts & \textbf{55.1} & 0.1\% & \textbf{51.5}  \\        
PointNeXt~\cite{qian2022pointnext}  & 0.01\%  &  44.7 & 20pts & 54.6 & 0.1\% & 48.9  \\
\rowcolor{green!10} \textbf{AADNet (PointNeXt)} & 0.01\%  &  \textbf{60.8} & 20pts & \textbf{62.5} & 0.1\% & \textbf{53.3}  \\
\specialrule{0em}{1pt}{1pt}
\toprule[3\arrayrulewidth]  
\end{tabular}
}
\end{center}
\caption{Quantitative comparisons under uniformly distributed sparse annotations.}
\label{tab:uniform}
\end{table}

In Fig.~\ref{fig:baseline_inhomo}, we compare the performance of the Baseline with AADNet under different sparsity (determined by the label rate) and inhomogeneity (determined by the number of labeled clusters) annotations. AADNet outperforms the Baseline in all 16 sparse labeling scenarios with different levels of difficulty. The superiority of AADNet is more significant at low labeling ratios since the negative effects caused by inhomogeneity are more severe at a lower label rate.

To compare with other alternatives in the non-uniform distribution sparse scenario, we reproduce four open-sourced methods. We choose the most inhomogeneous sparse labeling setting, \textit{i.e.}, $G = 1$, to demonstrate the ability to handle inhomogeneous annotations. Tab.~\ref{tab:non-uniform} illustrates that non-uniform annotation poses a significant hindrance to network learning. The comparison results demonstrate the effectiveness of AADNet with inhomogeneous annotations.

Even though AADNet targets arbitrarily distributed labeling, it outperforms previous methods under uniformly distributed labeling in Tab.~\ref{tab:uniform}.

\begin{figure}[t]
    \centering
    \includegraphics[width=0.9\linewidth]{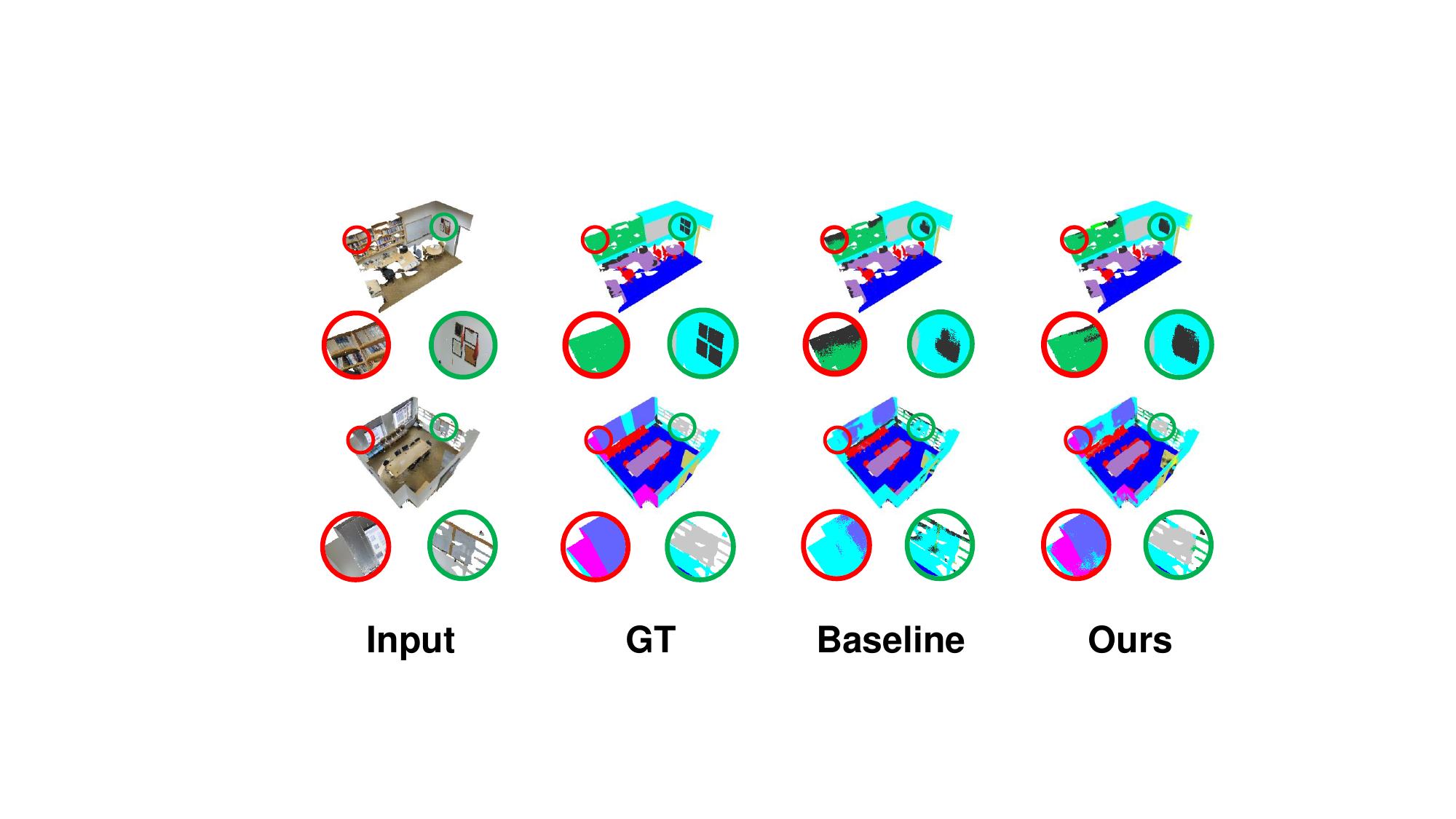}
    \caption{Qualitative comparisons on S3DIS under various label settings. The label settings are 10\% non-uniform (first row) and 0.01\% uniform sparse annotations (second row).}
    \label{fig:compare}
\end{figure}

In addition, we present a qualitative comparison of the method with the baseline in Fig.~\ref{fig:compare}. The baseline has a significant performance degradation on categories with variable forms, such as the bookcase with miscellaneous books and clutter on tables, and categories that are easily confused, such as wall, column, and board. Conversely, AADNet shows more complete predictions for head categories and more accurate predictions for long-tail categories.

\subsection{Ablation Study}
We perform extensive ablation experiments for the proposed method on S3DIS Area 5 to verify the effectiveness of each module. Baseline denotes the point cloud semantic segmentation network PointNeXt-L~\cite{qian2022pointnext} trained with the official default settings. The cluster number $G$ of inhomogeneous annotations in the ablation study is set to 1.

\noindent \textbf{Ablation on LaDS.} Tab.~\ref{tab:LaDS} demonstrates the segmentation performance with or without the label-first sampling. It can be observed that point-level label-first downsampling and voxel-level label-first downsampling promote the segmentation performance, respectively. However, voxel-level label-first downsampling may destroy the sampling diversity of resampling and therefore ignores part of the geometric topology. Hence, using both voxel-level and point-level label-first downsampling is suboptimal. The optimal performance is achieved by only using point-level label-first sampling (\emph{i.e.}, LaDS), which improves the label rate while avoiding compromising sampling diversity.

\begin{table}[t]
\begin{center}
\setlength{\tabcolsep}{8.7pt}
\resizebox{\linewidth}{!}{
\begin{tabular}{cc|cc|cc}
\toprule[3\arrayrulewidth]
\multicolumn{2}{c|}{Sampling Strategy}  & \multicolumn{2}{c|}{Homo. \tiny{($G=M$)}} & \multicolumn{2}{c}{Inhomo. \tiny{($G=1$)}} \\ 
Point-level & Voxel-level   & 0.01\%  & 0.1\%  & 1\% & 10\%              \\ 
\specialrule{0em}{1pt}{1pt}
\hline
\specialrule{0em}{1pt}{1pt}
 $\mathcal{S}\text{ampling}_R$ & $\mathcal{S}\text{ampling}_R$ & 44.7 & 64.3 & 35.9 & 54.1\\
 $\mathcal{S}\text{ampling}_R$ &$\mathcal{S}\text{ampling}_L$ & 45.6 & 65.0 & 36.4 & 54.0\\
$\mathcal{S}\text{ampling}_L$ & $\mathcal{S}\text{ampling}_L$ & 58.1 & 66.6 & 37.0 & 53.8\\
$\mathcal{S}\text{ampling}_L$ & $\mathcal{S}\text{ampling}_R$ & \textbf{58.4} & \textbf{67.0} & \textbf{37.8} & \textbf{54.8}\\
\specialrule{0em}{1pt}{1pt}
\toprule[3\arrayrulewidth]
\end{tabular}
}
\end{center}
\caption{Ablation study for LaDS on S3DIS Area 5.}
\label{tab:LaDS}
\end{table}

\begin{table}
\begin{center}
\begin{tabular}{l|cc|cc}
\toprule[3\arrayrulewidth]
\multicolumn{1}{c|}{\multirow{2}{*}{Setting}} & \multicolumn{2}{c|}{Homo. \tiny{($G=M$)}} & \multicolumn{2}{c}{Inhomo. \tiny{($G=1$)}} \\ 
\multicolumn{1}{c|}{}                         & 0.01\%  & 0.1\%  & 1\%              & 10\%              \\ 
\specialrule{0em}{1pt}{1pt}
\hline
\specialrule{0em}{1pt}{1pt}
 Baseline & 44.7 & 64.3 & 35.9 & 54.1\\
+ FixedW & 43.3 & 60.5 & 35.6 & 55.1\\
+ ADE & 48.2 & 64.7 & 36.3 & 55.0\\
+ MDE-ST & - & - & - & -\\
+ \textbf{MDE-AT} & \textbf{49.7} & \textbf{65.0} & \textbf{36.8} & \textbf{56.0}\\
\specialrule{0em}{1pt}{1pt}
\toprule[3\arrayrulewidth]
\end{tabular}
\end{center}
\caption{Ablation study for MDE-AT on S3DIS Area 5. '-' indicates that partial cross-entropy loss does not converge.}
\label{tab:MDE}
\end{table}

\begin{table}[htbp]
    \centering
    \setlength{\tabcolsep}{8.7pt}
    \resizebox{\linewidth}{!}{
    \begin{tabular}{c|ccc}
    \toprule[3\arrayrulewidth]
        Method & Labeling & Supervoxels & Pseudo-labels \\
        \specialrule{0em}{1pt}{1pt}
        \hline
        \specialrule{0em}{1pt}{1pt}
        Baseline & 44.7\% & 52.1\% & 53.0\% \\
        AADNet & 60.8\% \tiny{(+16.1\%)} & 61.7\% \tiny{(+9.7\%)} & 62.5\% \tiny{(+9.5\%)} \\
        \specialrule{0em}{1pt}{1pt}
    \toprule[3\arrayrulewidth]
    \end{tabular}
    }
    \caption{Validation on three inhomogeneously labeled scenarios at 0.01\% label rate.}
    \label{tab:val}
\end{table}

\begin{figure}[htbp]
  \centering
  \includegraphics[width=0.9\linewidth]{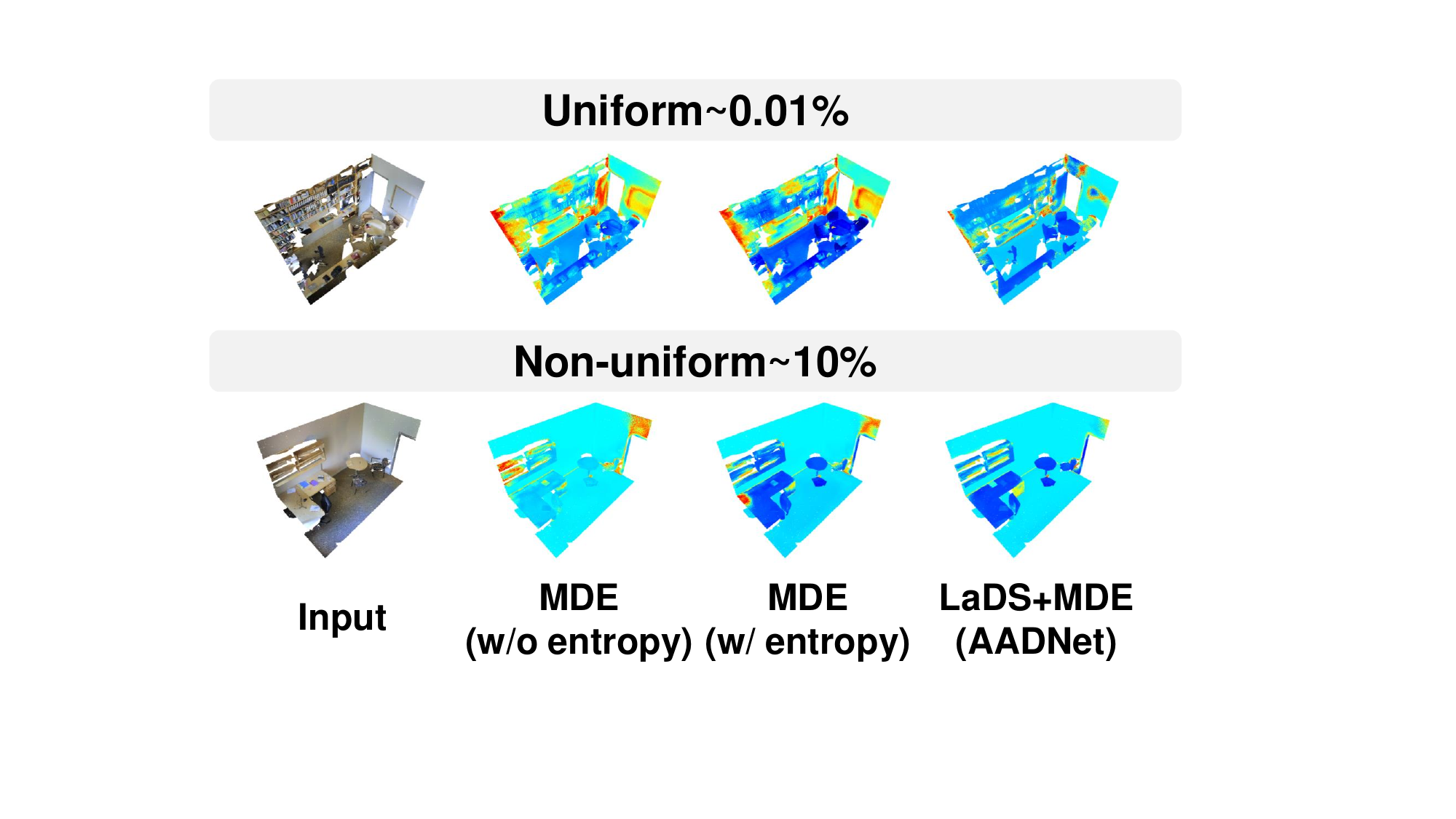}
  \caption{Shannon entropy visualization on predictions.}
  \label{fig:entropy}

\end{figure}

\noindent \textbf{Ablation on MDE-AT.} We validate several feasible gradient calibration functions in Tab.~\ref{tab:MDE}. FixedW sets the inverse of the local density of points as the gradient calibration function. ADE denotes the Additive Dynamic Entropy. MDE-ST and MDE-AT denote the multiplicative dynamic entropy with simultaneous training and asynchronous training, respectively. By observing the metrics, we have the following inferences: Fixed weights cannot adjust the weights dynamically for different point clouds. The additive calibration function is difficult to learn compared with the multiplicative one. Simultaneous training changes the learning objective of the partial cross-entropy loss function, leading to the partial cross-entropy loss function failing to converge.

\subsection{Further Analysis} \label{sec:further}
\noindent \textbf{Beyond simulation.} The simulated inhomogeneous annotations validate the effectiveness of labeling scenarios. We further report the performance under the other two scenarios in Tab.~\ref{tab:val}. With supervoxels or pseudo-labels strategies, AADNet still has significant improvements.

\noindent \textbf{Epistemic uncertainty.} Fig.~\ref{fig:entropy} visualize the entropy of predicted class probability. High-entropy areas are located in objects with variable forms and objects that are easily confused. By explicitly learning the entropy of the labeled points with asynchronous training, the entropy of these objects is significantly reduced, resulting in a reliable and robust semantic segmentation network. Finally, less epistemic uncertainty is achieved by combining LaDS and MDE-AT.

\section{Conclusion}
In this paper, we provide benchmarks for semantic segmentation with various distribution sparse annotations. By introducing the probability density function into the gradient sampling approximation analysis, we reveal the impact of sparse annotation distributions in the manual labeling process. Based on gradient analysis, we propose an adaptive annotation distribution network, consisting of the label-aware downsampling strategy and the multiplicative dynamic entropy for asynchronous training. AADNet achieves robust learning in different labeling scenarios. We expect that our work can provide novel research insights for the weakly supervised point cloud semantic segmentation community.

\section*{Acknowledgements}
This work was supported in part by National Science and Technology Major Project (2024ZD01NL00101), and in part by Guangdong Provincial Key Laboratory of Ultra High Definition Immersive Media Technology (2024B1212010006).

\bibliography{Zhiyi}


\end{document}